\documentclass{article}
\usepackage{spconf}
\usepackage{amsmath}
\usepackage{graphicx}
\usepackage{subcaption}
\usepackage{multirow}
\usepackage{tabularx}
\usepackage{makecell}
\usepackage{caption} 

\usepackage{booktabs}
\usepackage{adjustbox}
\usepackage{lipsum}
\usepackage{verbatim}
\usepackage{url}
\usepackage{amssymb}

\title{Automated Speaking Assessment of Conversation Tests with Novel Graph-based Modeling on Spoken Response Coherence}
\name{Jiun-Ting Li$^{1,2}$\sthanks{The first author performed the work while at National Taiwan Normal University.}, Bi-Cheng Yan$^1$, Tien-Hong Lo$^1$, Yi-Cheng Wang$^1$, Yung-Chang Hsu$^2$ and Berlin Chen$^1$}
\address{
  $^1$Dept. Computer Science and Information Engineering, National Taiwan Normal University, Taiwan \\
  $^2$EZAI, Taiwan\\
  \texttt{\{jtlee, bicheng, teinhonglo, yichengwang, berlin\}@ntnu.edu.tw}
}

\begin{document}
%
\maketitle
\begin{abstract}
Automated speaking assessment in conversation tests (ASAC) aims to evaluate the overall speaking proficiency of an L2 (second-language) speaker in a setting where an interlocutor interacts with one or more candidates. Although prior ASAC approaches have shown promising performance on their respective datasets, there is still a dearth of research specifically focused on incorporating the coherence of the logical flow within a conversation into the grading model. To address this critical challenge, we propose a hierarchical graph model that aptly incorporates both broad inter-response interactions (e.g., discourse relations) and nuanced semantic information (e.g., semantic words and speaker intents), which is subsequently fused with contextual information for the final prediction. Extensive experimental results on the NICT-JLE benchmark dataset suggest that our proposed modeling approach can yield considerable improvements in prediction accuracy with respect to various assessment metrics, as compared to some strong baselines. This also sheds light on the importance of investigating coherence-related facets of spoken responses in ASAC.

\end{abstract}

\section{Introduction}


\begin{figure}[h!]
  \centering
  \includegraphics[width=\linewidth]{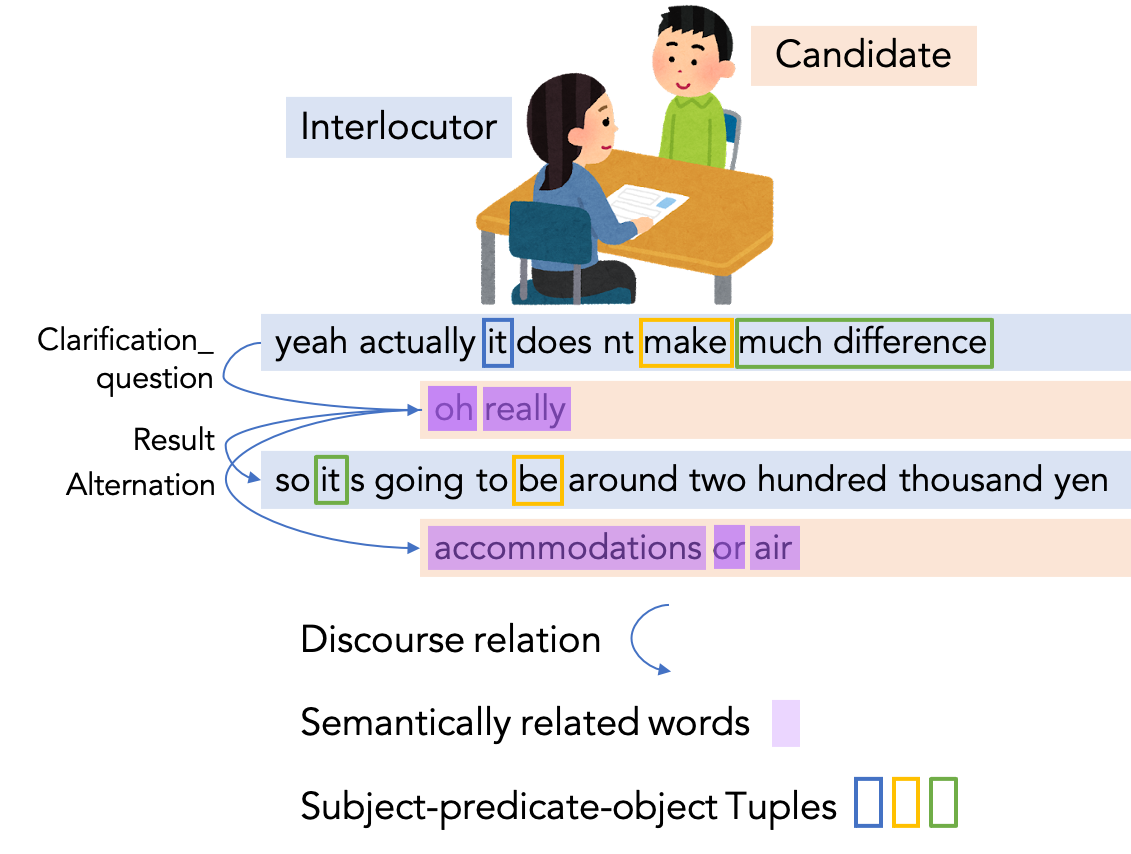}
  \caption{An illustration of a partial conversation test sample in the NICT JLE \cite{izumi_nict_2004} dataset. Each sentence is the response to the conversation. The conversation test is held under specific topics and with the participation of an interlocutor and a candidate.}
  \label{fig:conversation_test}
\end{figure}


The rising demand for English-speaking skills in business and academic settings has sparked a fascination with automated systems for assessing and teaching spoken languages. As such, the technology of computer-assisted language learning (CALL) has emerged as a promising solution, offering instant and effective feedback to complement traditional curricula. CALL-related research is not only of sufficient academic interest to multidisciplinary communities of scientists and engineers, but has making inroads into commercial products with significant education market values \cite{222431475_Embodied_conversational_agents_in_Computer_Assisted_Language_Learning}. Practical examples include classroom solutions \cite{kelly2020soapbox} to humanoid assist teachers or rigorous assessments \cite{xu2020linguaskill}, which address the demand of businesses and academic institutes to ensure employees and students to exceed a fundamental level of English proficiency, thereby enabling them to thrive and make meaningful contributions within their respective environments. On a separate front, standardized measurements, such as the common European framework of reference (CEFR) \cite{2001_cefr} scoring, are widely accepted in languages like English, German, French, and others. 

Previous research on automated speaking assessment (ASA) has explored various modeling frameworks to assess spoken content, to name a few, multi-task framework \cite{craighead_etal_2020_investigating} and multimodal architecture \cite{mcknight23_slate}, which aim to capture the complementarity of linguistic and acoustic aspects. Previous research also made efforts to investigate representations related to various facets of language proficiency, including hand-crafted features \cite{chen2018automated}, self-supervised learning (SSL) representations \cite{mcknight23_slate, banno23_slate} and others, for seamlessly retaining the interpretability and the controllability as well as promoting the stability and performance of ASA systems. Albeit the recent advancements of modeling frameworks for proficiency testing, most of them simply treat spoken responses from a perspective of sequential input, which may overlook the dynamic human-to-human or human-to-machine interactions inherent in conversation tests. For example, a human-to-human conservation scenario illustrated in Figure \ref{fig:conversation_test} demonstrates rich hierarchical contexts in both inter- and intra-responses, which are hard to be adequately captured with sequential models alone \cite{ijcai2021p0524}.

These challenges are evident when trying to extend across conversational turns at a broader scope. In the course of communication, coherence is in charge of the logical transmission that the language learners negotiate the meaning of conversations with the interlocutor, or the interlocuter seeks to elicit the best responses from candidates. It can be delineated across two distinct levels, each capturing the essence of speaker intent but operating at different linguistic granularities. At the macro level, we have the intentional structures that involve a series of dialogue acts \cite{1962:austin}, primarily carried out through speech, serving as markers for transitions within a conversation at the sentence level. At the micro level, the finer-grained knowledge units within a response demonstrate the actual action \cite{10.1145/3539597.3570415}. Both of them figure prominently to be the vital elements in conversation \cite{cervone18_interspeech}, revealing a gap in investigating them within interactable assessments.

Additionally, previous methods assess the candidate alone, it is imperative to consider the fact that the candidate's spoken content can vary widely. Thus, the incorporation of the interlocutor's spoken content into modeling retains overall conversation fluency and provides a more comprehensive perspective on completeness.

\begin{figure*}[!]
  \centering
  \includegraphics[width=\linewidth]{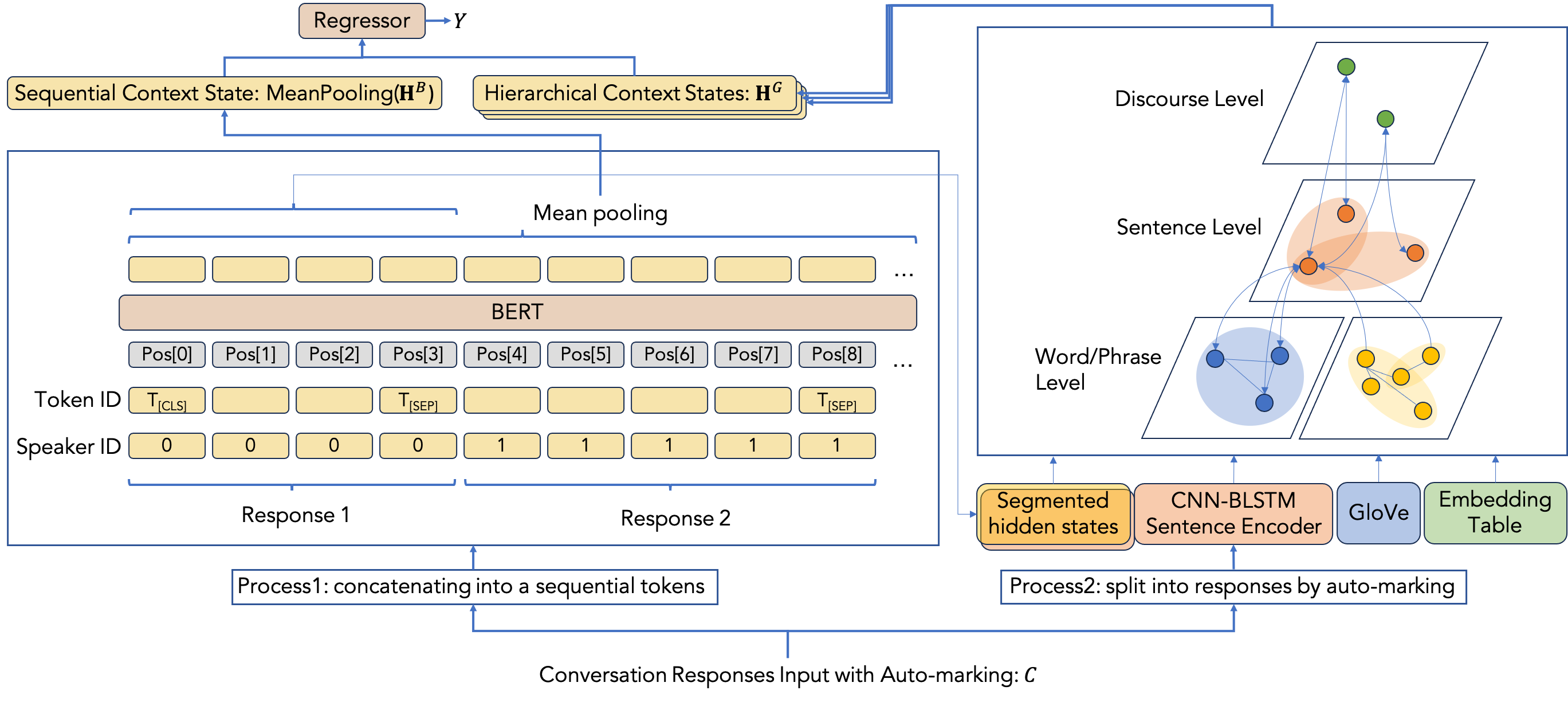}
  \caption{This illustrates the ASAC grading model framework. From bottom to top, it includes: (1) Two processing stages for conversational spoken content—concatenating and splitting—to prepare model inputs. (2) The left module is a contextualized encoder for sequential input, and the right module models hierarchical contexts in inter- and intra-responses with hierarchical levels and a bottom-to-top propagation path. Hierarchical context in conversation data is proposed to implement coherence. At the response level (intra-responses), it aggregates semantic information from semantically related words and intents from the SPO tuple. This response information then propagates to the discourse level for inter-response interaction. (3) The graph-based representation $\textbf{H}^{G}$ is fused with the mean pooled embedding of $\textbf{H}^{B}$ which is derived from the sequential model, to form the final decision $Y$.}
  \label{fig.methodology.hierarchical_context}
  \vspace{-0.5cm}
\end{figure*}

To address the foregoing limitations, we propose an innovative approach: a hierarchical schematic graph modeling method for conversation, as illustrated in Figure \ref{fig.methodology.hierarchical_context}. This method converts spoken content in a conversation to a graph and dissects it into multiple layers, spanning from individual words to broader discourse structures. The semantic information within intra-responses is aggregated into the response level from two decoupled granularities: the underlying semantic context and the actions encapsulated within responses. In the former granularity, it emphasizes the semantically related words, which promotes the performance of grading models \cite{li23_slate}. In the latter granularity, actions utilize the pattern subject-predicate-object (SPO) triplets in responses \cite{10.1145/3539597.3570415}, which express the fine-grained speaker's intent at the in-depth view of intra-responses. Then, the response's information propagates to the coarser-grained discourse level to refine structural knowledge of coherence, fusing it into the final decision. As evident in our experimental findings, effectively capturing coherence within conversations can significantly enhance the grading model's ability to identify pivotal content, thereby succinctly facilitating more precise proficiency assessment.

In summary, our contributions in this paper are at least three-fold: (1) We present a method of enhanced hierarchical graph modeling to obtain more expressive representations pertaining to coherence information, benefiting from message propagation. (2) We shed light on the way to subsuming hierarchical context under the framework of the grading model. (3) Code for model and data preprocessing is available at \url{https://github.com/a2d8a4v/HierarchicalContextASA/}.

\section{Related Work}

\subsection{Automated Speaking Assessment}

Research on L2 ASA iconicly adopted feature-based approaches to score prediction \cite{chen2018automated, zechner2019automated}. Prior arts commonly employed hand-crafted features derived from the involvement of automatic speech recognition systems. These are related to various facets of language proficiency, such as fluency \cite{strik1999automatic, preciado2018speaker}, pronunciation \cite{10.1007/978-3-030-50729-9_48, 5700836}, prosody \cite{coutinho2016assessing}, and rhythm \cite{kyriakopoulos19_interspeech}, applied in many automatic scoring tasks and become the building block in this field.

Recently, the rise of deep learning-based paradigms, such as SSL models: BERT \cite{devlin_etal_2019_bert} or Wav2vec2.0 \cite{baevski2020wav2vec}, fill the gap that hand-designed features may inevitably leave out some salient information relevant to ASA. These models have showcased excellent capability to represent contextual information, and are successfully integrated into various language assessment tasks. Despite the decent efficacy of the modeling methods, the exploration of hierarchical context has aroused increasing attention due to structural knowledge is largely dismissed in spoken content modeling for ASA. As such, we in this paper further investigate the hierarchical context in the aspect of coherence.

\subsection{Knowledge Graph-enhanced Modeling}
Knowledge graphs (KGs) explicitly store rich factual knowledge. KGs can enhance the language model (LM) by providing external knowledge for inference and interpretability. Such a measure has proven effective in many tasks, such as entity retrieval \cite{10.1145/3539597.3570415}, reranking \cite{chiu2021crossutterance}, conversation summarization \cite{ijcai2021p0524} and many others. It is reliable for integrating human-defined knowledge and imitating structural objects like in the real world. In this work, we follow a roadmap that KG is an enhanced knowledge to fuse in the final decision.


\section{Methodology}

\subsection{Task Formulation}

We tackle the ASAC task with the following processing flow. Given $C$ is the entire spoken content in a conversation test, $C = \{u_{0},  u_{1}, ... u_{|C|}\}$, with $|C|$ responses, which consists of a sequence of speech utterances intertwined with interlocutor’s and subsequently candidate’s responses. Our objective is to design and develop a function that can predict the final holistic proficiency $Y$ considering the sequential contextual information and hierarchical context inherent in $|C|$.

\subsection{Overall Framework}

We present a conversation grading model that is equipped with an essential LM and an enhanced hierarchical graph module (EHGM), defined as follows:
\vspace{-0.5cm}

\begin{align}
  \textbf{H}^{B} &= \text{LM}(\text{process}_{1}(C)), \\
  \textbf{H}^{G} &= \text{EHGM}(\text{process}_{2}(C), \textbf{H}^{B}), \\
  \hat{Y} &= \text{Regressor}(\text{MeanPooling}(\textbf{H}^{B}), \textbf{H}^{G}).
\end{align}
\vspace{-0.5cm}

LM$(\cdot)$ captures contextualized information from the conversation's sequential input, while EHGM$(\cdot)$ captures hierarchical conversation context, spanning multiple levels: discourse relations at the highest level, followed by sentence-level relations, and finally word or phrase level relations. The propagation path moves from lower to higher levels. It gathers information from word and phrase nodes to update sentence representations and augment discourse relation representations. A regressor combines these refined embeddings to predict final holistic proficiency.

\subsection{Structured Graph Construction}

This section outlines the construction of graphs to depict hierarchical contexts within conversation data. Spoken content in conversation is converted into hierarchical layers, transitioning from word or phrase level to sentence level and culminating at the highest discourse level. These assemble the semantic content with two decoupled granularities: the actions encapsulated within responses and the underlying semantic context. Sentence nodes connect to build two separate graphs. Similarly, discourse nodes link to sentence nodes, constructing a sentence-aware discourse relation graph. Inspired by \cite{du-etal-2023-structure}, each subgraph employs a global node to unidirectionally summarize the exchanged information in the other nodes.

At various levels of granularity within the structured graph construction process, we define graphs $G^{c}$, $G^{a}$, and $G^{d}$ for semantically related words, actions, and discourse relations, respectively. Each with heterogeneous nodes spanning different linguistic levels. $G^{c}$ represents the word or phrase level, where $V^{c}$ includes word nodes, response nodes, and global nodes, and $E^{c}$ denotes the connecting edges. Similarly, $G^{a}$ represents speaker intents, incorporating nodes such as subject, predicate, object, intent nodes, response nodes, and global nodes within $V^{a}$, while $E^{a}$ signifies the edges linking these diverse nodes. Furthermore, $G^{d}$ encompasses response nodes, discourse nodes, and global nodes within $V^{d}$, with connections established through parsed relation links denoted by $E^{d}$.

\noindent \textbf{Semantically Related Words.} It aims to enhance the high-importance words existing in responses. Modeling semantically related words succeeds in natural language processing tasks \cite{li23_slate, wang-etal-2020-heterogeneous, phan-etal-2022-hetergraphlongsum} but has yet to be applied in ASAC. We initialize word nodes using GloVe \cite{pennington-etal-2014-glove} and employ a residual multilayer perceptron (MLP) layer for response nodes. The initialization process is depicted as follows:
\vspace{-0.5cm}

\begin{align}
  \textbf{H}^{s}_0 &= \text{MLP}(\textbf{H}^{B}[p_{start}:p_{end}] \odot \textbf{H}^{s}_{ngm}) + \textbf{H}^{s}_{ngm},
\end{align}
\vspace{-0.5cm}

\noindent where $p_{start}$ and $p_{end}$ are the position indexes of a response in a sequential conversation input, $\textbf{H}^{B}$ is the last hidden state from the contextualized encoder, $\textbf{H}^{s}_{ngm}$ is the $\textit{n}$-gram embedding for that response, and $\odot$ represents the concatenation operation at the axis of dimension. We follow the idea in \cite{wang-etal-2020-heterogeneous} to generate $\textbf{H}^{s}_{ngm}$ by a CNN-BLSTM encoder.

\noindent \textbf{Actions.} It aims to identify the intra-response lexical intent at the word or phrase level. We utilize an open information extraction toolkit \cite{Stanovsky2018SupervisedOI} to filter the knowledge units (SPO triplets) from each response in the conversation. In $G^{a}$, the intent node is to absorb a tuple of SPO nodes to represent a speaker's intent in a response. Many intent nodes then connect to their corresponding sentence nodes. The node initializer follows \cite{10.1145/3539597.3570415}.

\noindent \textbf{Discourse Relations.} This component aims to identify the inter-response speaker intent in the logical flow of transmission between responses. Ground-truth discourse relations are not provided, so we exploit a pre-trained state-of-the-art model \cite{chi-rudnicky-2022-structured} to obtain the links of discourse relations among continuous responses in conversations. We follow the ideas in \cite{ijcai2021p0524} to perform a Levi graph transformation \cite{gross2013handbook}, converting these relations and sentences into nodes, with sentence-relation links as directed edges in the graph. We then initialize the discourse node embeddings with an embedding table. This approach allows us to model discourse relations explicitly and update the utterance and discourse vertices simultaneously.

\subsection{Encoders}

Given a sequential conversation input and multiple graphs, we utilize two main kinds of encoders to capture and represent information.

\noindent \textbf{Contextualized Encoder.} The goal of the contextualized encoder is to capture contextual information from the conversation text. To compete with \cite{mcknight23_slate}, we also utilize the RoBERTa model \cite{liu2019roberta} as the LM,
However, conversations often exceed the maximum input length that RoBERTa can process in a single pass. To address this limitation, we adapted RoBERTa to handle conversations with a maximum length of 1,600 tokens using a sliding window approach \cite{beltagy2020longformer}. Additionally, we incorporate segmental indexical embeddings to distinguish characters in the conversation. The contextualized encoder outputs a last hidden state representing the conversation for downstream modules.

\noindent \textbf{Graph Encoders.} Building upon the contextualized encoder, graph encoders aim to learn expressive representations of nodes in the constructed graphs. After graph construction and node embedding initialization, graph encoders possess a graph attention network (GAT) \cite{veličković2018graph} backbone to capture the data heterogeneity under the scheme of the multi-head attention head. The GAT follows a neighborhood aggregation scheme, simplifiable into two key functions: \verb|combine| and \verb|aggregate|, in other words, computes the node representations by recursively encapsulating node features from local neighborhoods. For more details about the operation process, please refer to \cite{wang-etal-2020-heterogeneous}.

The residual connection is applied to sidestep the gradient vanishing problem. After each graph attention layer, we utilize a position-wise feed-forward layer composed of two linear convolutional transformations.

\subsection{Regressor}

The regressor module utilizes several sources of embeddings, including in $\textbf{H}^{inventory}=\{\textbf{H}^{B}_{mp},\textbf{H}^{G^{c}},\textbf{H}^{G^{a}},\textbf{H}^{G^{d}}\}$, $\textbf{H}^{B}_{mp}$ denotes mean pooled embedding of $\textbf{H}^{B}$, to predict the holistic proficiency score $\hat{Y}$. The mechanisms in the regressor are implemented as the following formula:
\vspace{-0.5cm}

\begin{align}
    \textbf{H}^{\text{combo}}_{(m,k)} &= \textbf{H}_{m} \odot \textbf{H}_{k} \in \mathbb{R}^{2D_{\textbf{H}}}, \\
    \textbf{H}^{\text{head}}_{(m,k,l)} &= \text{ReLU}(A_k(\textbf{H}^{\text{combo}}_{(m,k)})) \in \mathbb{R}^{D_{\textbf{H}}}, \\
    \textbf{H}_{\text{combo}} &= \odot_{\substack{m,k \in \{1, \ldots, N_{\text{combo}}\} \\ l \in \{1, \ldots, N_{\text{h}}\}}} \textbf{H}^{\text{head}}_{(m,k,l)} \in \mathbb{R}^{D_{\textbf{H}} N_{\text{h}} N_{\text{combo}}}, \\
    \hat{Y} &= \text{Linear}(\textbf{H}_{\text{combo}}),
\end{align}
\vspace{-0.5cm}

\noindent where $D_{\textbf{H}}$ is the dimension of hidden states, $N_{\text{h}}$ is the number of heads. $\textbf{H}^{\text{combo}}_{(m,k)}$ represents the combined embedding of pair $(m,k)$ from $\textbf{H}^{inventory}$, $m \neq k$, and $N_{\text{combo}} = \frac{1}{2}N^{\text{Reg.}}_{\text{inputs}}(N^{\text{Reg.}}_{\text{inputs}}-1)$, $N^{\text{Reg.}}_{\text{inputs}}$ is the number of inputs for the regressor. $\textbf{H}^{\text{head}}_{(m,k,l)}$ denotes the output of the $l$-th attention head, $\textbf{H}_{\text{combo}}$ is the concatenated output of all attention heads, and $\hat{Y}$ represents the predicted proficiency score obtained from the linear function applied to $\textbf{H}_{\text{combo}}$.

\subsection{Optimization}

\noindent \textbf{Reweighted Loss Function.} During the training process, we seek to minimize the weighted mean squared error (MSE) loss between the predicted and target holistic score. The strategy is adopted from \cite{li23_slate}.

\noindent \textbf{Posttraining Strategy on Multiple Auxiliary Tasks.} This strategy aims to optimize the initial parameters of the contextualized encoder. Inspired by \cite{craighead_etal_2020_investigating}, we facilitate the LM's ability to learn multi-aspects of assessment knowledge and promote prediction accuracy on the main scoring task. It follows the training stages in \cite{banno-matassoni-2022-cross}. In the first three epochs, the model undergoes posttraining on the EFCAMDAT dataset \cite{geertzen2013automatic}, which is rich in full CEFR labels (A1-C2). Subsequently, the model is trained on the NICT JLE corpus in the next three epochs. This strategy is applied to optimize the initial parameters of the contextualized encoder.

\section{Experiments}

\subsection{Experimental Setup}

The conversation grading model is trained using the Adam optimizer \cite{kingma2014adam} with a batch size of 64 and gradient accumulation steps set to two. To address model uncertainty \cite{wu20n_interspeech}, each experiment is repeated five times with an exponentially decaying learning rate initialized at various values and a decay factor of 0.85 per epoch. Training stops if the validation loss does not decrease for four consecutive epochs, and the last epoch is used as the best checkpoint.

Model performance evaluation metrics include root-mean-square error (RMSE), Pearson's correlation coefficient (PCC), and margin accuracy within 0.5 and 1.0, which is the percent of predictions that are within a half or one score from the reference score \cite{craighead_etal_2020_investigating}. The margin accuracy metric is assessed at both the micro and macro levels (among samples / CEFR groups) to provide a comprehensive understanding of model accuracy.

\begin{table}[t]
\centering
\small
\caption{Results of the grading model: we reproduce \cite{mcknight23_slate} BERT is the baseline model. The proposed measures are denoted as C, D, and A, representing semantically related words, discourse relations, and actions within responses, respectively. Micro accuracy and macro accuracy are abbreviated as Acc. and M. Acc., respectively \cite{craighead_etal_2020_investigating}.}
\label{tab:experimentalresults.mainresults}
\begin{tabularx}{\linewidth}{@{}>{\raggedright\arraybackslash}m{0.9cm}|@{\hskip 5pt}*{6}{>{\centering\arraybackslash}X@{\hskip 2pt}}@{}}
\toprule
\multirow{2}{*}{\textbf{Model}} & \multirow{2}{*}{\textbf{RMSE}$\downarrow$} & \multirow{2}{*}{\textbf{PCC}$\uparrow$} & \multicolumn{2}{c}{\textbf{Acc.}$\uparrow$} & \multicolumn{2}{c}{\textbf{M. Acc.}$\uparrow$} \\
 & & & $\leq$0.5 & $\leq$1.0 & $\leq$0.5 & $\leq$1.0 \\ 
\midrule
{\footnotesize \makecell[l]{BERT\\\cite{mcknight23_slate}}} & {\footnotesize \makecell{0.656 \\ (0.034)}}& {\footnotesize \makecell{0.273 \\ (0.075)}}& {\footnotesize \makecell{61.812 \\ (2.945)}}& {\footnotesize \makecell{85.890 \\ (1.498)}}& {\footnotesize \makecell{29.687 \\ (2.813)}}& {\footnotesize \makecell{55.340 \\ (3.401)}}\\
{\footnotesize \makecell[l]{\hspace{1mm}+C\\\cite{li23_slate}}} & {\footnotesize \makecell{0.539 \\ (0.022)}}& {\footnotesize \makecell{0.711 \\ (0.036)}}& {\footnotesize \makecell{64.234 \\ (0.223)}}& {\footnotesize \makecell{93.677 \\ (0.820)}}& {\footnotesize \makecell{47.024 \\ (4.811)}}& {\footnotesize \makecell{81.489 \\ (8.288)}}\\
{\footnotesize \hspace{1mm}+D} & {\footnotesize \makecell{0.520 \\ (0.005)}}& {\footnotesize \makecell{0.729 \\ (0.001)}}& {\footnotesize \makecell{65.961 \\ (0.518)}}& {\footnotesize \makecell{94.763 \\ (0.189)}}& {\footnotesize \makecell{42.037 \\ (0.357)}}& {\footnotesize \makecell{84.671 \\ (0.160)}}\\
{\footnotesize \hspace{1mm}+CD} & {\footnotesize \makecell{0.502 \\ (0.001)}}& {\footnotesize \makecell{0.751 \\ (0.000)}}& {\footnotesize \makecell{67.437 \\ (0.378)}}& {\footnotesize \makecell{96.323 \\ (0.167)}}& {\footnotesize \makecell{57.731 \\ (0.405)}}& {\footnotesize \makecell{92.936 \\ (0.141)}}\\
{\footnotesize \hspace{1mm}+CDA} & {\footnotesize \makecell{0.507 \\ (0.001)}}& {\footnotesize \makecell{0.755 \\ (0.000)}}& {\footnotesize \makecell{66.323 \\ (0.346)}}& {\footnotesize \makecell{96.407 \\ (0.056)}}& {\footnotesize \makecell{55.654 \\ (0.901)}}& {\footnotesize \makecell{91.968 \\ (0.066)}}\\ \midrule
{\footnotesize C} & {\footnotesize \makecell{0.525 \\ (0.002)}}& {\footnotesize \makecell{0.724 \\ (0.001)}}& {\footnotesize \makecell{65.376 \\ (0.287)}}& {\footnotesize \makecell{93.928 \\ (0.189)}}& {\footnotesize \makecell{49.310 \\ (0.218)}}& {\footnotesize \makecell{82.170 \\ (0.257)}}\\
{\footnotesize D} & {\footnotesize \makecell{0.567 \\ (0.068)}} & {\footnotesize \makecell{0.643 \\ (0.142)}} & {\footnotesize \makecell{63.760 \\ (3.434)}} & {\footnotesize \makecell{92.618 \\ (3.835)}} & {\footnotesize \makecell{38.780 \\ (6.691)}} & {\footnotesize \makecell{75.889 \\ (14.771)}} \\
{\footnotesize C+D} & {\footnotesize \makecell{0.502 \\ (0.001)}} & {\footnotesize \makecell{0.745 \\ (0.001)}} & {\footnotesize \makecell{67.660 \\ (0.690)}} & {\footnotesize \makecell{95.933 \\ (0.056)}} & {\footnotesize \makecell{47.533 \\ (0.880)}} & {\footnotesize \makecell{89.485 \\ (0.111)}} \\ 
{\footnotesize C+D+A} & {\footnotesize \makecell{0.501 \\ (0.001)}}& {\footnotesize \makecell{0.745 \\ (0.001)}}& {\footnotesize \makecell{67.584 \\ (0.752)}}& {\footnotesize \makecell{95.926 \\ (0.060)}}& {\footnotesize \makecell{47.452 \\ (0.967)}}& {\footnotesize \makecell{89.493 \\ (0.122)}}\\ \bottomrule
\end{tabularx}
\vspace{-0.5cm}
\end{table}

\subsection{Corpus}

The NICT JLE corpus \cite{izumi_nict_2004} comprises English oral conversations involving 1,281 Japanese individuals and 20 American native speakers, totaling two million human-annotated words. However, only text-annotated transcriptions are accessible, and ASR errors are absent. The conversations cover various topics and feature non-fixed prompt questions, with candidates expected to respond promptly. Each conversation is assigned a Standard Speaking Test (SST) score ranging from one to nine, which can be mapped to CEFR levels. For more information on data preprocessing, dataset distribution, and the SST-CEFR mapping table, please refer to \cite{li23_slate}.

The corpus includes detailed annotations for spoken content, covering aspects such as auto-marking of each response, non-fluency in speech, and instances where the interviewee's speech overlaps with the interviewer's \cite{tanimura2005learners, izumi2005error} \footnote{\url{https://alaginrc.nict.go.jp/nict_jle/src/taglist.pdf} describes the function of each tag. \\ \url{https://alaginrc.nict.go.jp/nict_jle/src/readme_transcription.pdf} accounts for the meaning of each grammar error tag with a succinct example.}. These annotations support detailed studies across different English proficiency levels. For our experiments, we preprocessed some annotations to simplify complexity and avoid overlap issues, which were minimal, and left detailed exploration for future work.

The human-annotated spoken content is pre-tokenized via the UDPipe toolkit \cite{straka-strakova-2017-tokenizing}, in which the LM is trained on a vast amount of web-available media text content. UDPipe also provides part-of-speech and morphosyntactic tagging, enriching the information used for post-training strategies on the initial parameters of the contextualized encoder. Filled pauses were temporarily removed before pre-tokenizing and added back afterward to avoid influencing the correctness of pre-tokenizing and tagging.

\subsection{Main Results}

In this section, we present the performance of our proposed methods and discuss the key findings from our experiments. Table \ref{tab:experimentalresults.mainresults} provides a comprehensive comparison of our approaches with a reference model, BERT \cite{mcknight23_slate}, which primarily utilizes contextual information to assess proficiency holistically. We utilize the contextualized encoder from \cite{mcknight23_slate} as the baseline model. Due to constraints, we only have access to the annotated transcriptions of the NICT JLE corpus, leading to the omission of the acoustic part. Additionally, there are differences in the maximum position input size.

\begin{figure}[t]
  \centering
  \includegraphics[width=\linewidth]{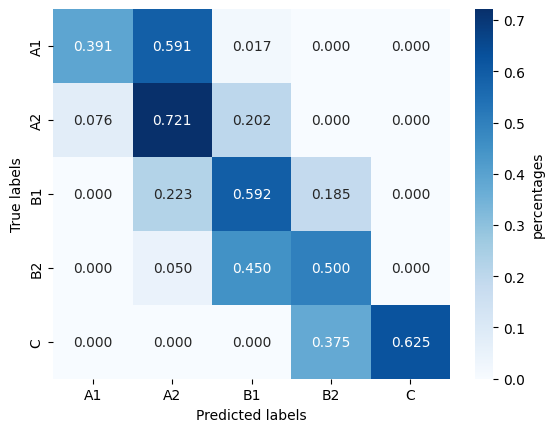}
  \caption{This illustration is obtained from the evaluation results of our proposed mothed (BERT+CDA). The confusion matrix depicts the percentage of prediction on the right position as targets in CEFR levels.}
  \label{fig:experimentalresults.allmethodfused.confusion_matrix}
  \vspace{-0.5cm}
\end{figure}

\begin{figure}[t]
  \centering
  \includegraphics[width=\linewidth]{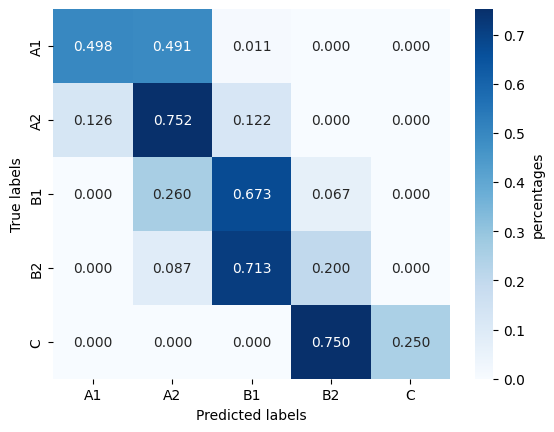}
  \caption{This illustration is obtained from the evaluation results of our proposed mothed (C+D+A).}
  \label{fig:experimentalresults.allhierarchical_context.confusion_matrix}
  \vspace{-0.5cm}
\end{figure}

Our proposed method demonstrates a notable overall improvement across all metrics. This outcome suggests that the enhanced graph modeling, incorporated into our approach, aids the grading model in effectively emphasizing the hierarchical context and speaker intents, outperforming the $\text{BERT}$ model with stable scores in other metrics. We further visualize these improvements in Figure \ref{fig:experimentalresults.allmethodfused.confusion_matrix}, highlighting that most scores align with the correct positions, albeit with a slight tendency to under-score.

\subsection{Ablation Studies}

In our quest to delve into the factors impacting the performance of the grading model, we conducted a series of ablation studies. These experiments were designed to dissect the influence of the hierarchical context on our model's effectiveness.

A critical component of our approach is the introduction of hierarchical context to enhance the LM on grading scores. Specifically, the components +CD and +CDA positively affect performance, indicating that both inter- and intra-response interactional information contributes significantly to conversational assessment. We then analyzed these components in detail. Method +C follows the ideas in \cite{li23_slate}, constructing semantically related information from the word level to the paragraph level, though it lacks sentence-level information. In our work, we build semantically related information at the sentence level, allowing connection with other proposed components. This method also shows greater performance compared to a pure BERT model, highlighting the importance of emphasized words. Method +D also has a performance improvement, showing the necessity of discourse relation to play a vital role in modeling the spoken response coherence. Conversely, while the addition of component A resulted in minor performance degradation, this observation highlights the complexity of effectively integrating intra-response speaker intent. This suggests that the current method of incorporating component A may need refinement. Future work will focus on developing more sophisticated techniques for selecting and integrating intra-response speaker intent to enhance overall model performance.

Furthermore, our evaluation results show that even with only hierarchical context components (C, D, C+D, C+D+A), the conversation grading model performs nearly as well as the graph-enhanced BERT model. However, graph modeling is limited by predefined structural knowledge, whereas LM can learn from unfactual knowledge. Figure \ref{fig:experimentalresults.allhierarchical_context.confusion_matrix}  illustrates that the model using only hierarchical context tends to predict one lower group than the graph-enhanced BERT model. This suggests that both components are indispensable in modeling conversation data.

\section{Conclusions}

In this paper, we explored the use of hierarchical graph modeling for conversation tests, which amalgamates hierarchical context, including semantically related words, actions, and discourse relations. A series of empirical experiments have revealed the efficacy of our proposed methods and the practical potential of extending the current ASAC framework to accommodate the modeling of spoken response coherence in conversation tests.

\noindent \textbf{Limitations and Future Work.} While this work focuses on investigating spoken response coherence in conversation tests, it does not explore other natural language processing tasks such as conversation summarization. Additionally, factors such as overlapping responses, error propagation from ASR systems, and error recognition by discourse-parsing models were not addressed, which are left for future exploration. In future work, we plan to model our task using continuous graph turns, incorporating the history of conversation turns to create a more realistic conversation test environment.

\bibliographystyle{IEEEbib}
\bibliography{strings,refs}

\end{document}